\documentclass[10pt,twocolumn,letterpaper]{article}

\usepackage{cvpr}              

\usepackage{amsmath,amssymb,amsfonts} 
\DeclareMathOperator*{\argmax}{arg\,max}

\usepackage{booktabs} 
\usepackage{multirow} 
\usepackage{makecell} 
\usepackage{array} 
\usepackage{colortbl} 
\usepackage{enumitem} 
\newcommand{\hl}{\cellcolor{blue!15}}  
\newcommand{\gain}[1]{\textcolor{green!30!black}{#1}}  
\usepackage{algorithm}
\usepackage{algpseudocode}

\definecolor{cvprblue}{rgb}{0.21,0.49,0.74}
\usepackage[pagebackref,breaklinks,colorlinks,allcolors=cvprblue]{hyperref}

\def\paperID{1803} 
\def\confName{CVPR}
\def\confYear{2026}

\title{Wavelet-based Frame Selection by Detecting Semantic Boundary for Long Video Understanding}

\author{Wang Chen$^{1}$, Yuhui Zeng$^{1}$, Yongdong Luo$^{1}$, Tianyu Xie$^{1}$\\
Luojun Lin$^{2}$, Jiayi Ji$^{1}$, Yan Zhang$^{1}$, Xiawu Zheng$^{1}$\thanks{Corresponding author: zhengxiawu@xmu.edu.cn.}\\
$^{1}$Key Laboratory of Multimedia Trusted Perception and Efficient Computing,\\
Ministry of Education of China, Xiamen University\\
$^{2}$College of Computer and Data Science, Fuzhou University
}

\begin{document}
\maketitle

\begin{abstract}
Frame selection is crucial due to high frame redundancy and limited context windows when applying Large Vision-Language Models (LVLMs) to long videos.
Current methods typically select frames with high relevance to a given query, resulting in a disjointed set of frames that disregard the narrative structure of videos.
In this paper, we introduce \textbf{W}avelet-based \textbf{F}rame \textbf{S}election by Detecting \textbf{S}emantic \textbf{B}oundary (\textbf{WFS-SB}), a training-free framework that presents a new perspective: effective video understanding hinges not only on high relevance but, more importantly, on capturing semantic shifts—pivotal moments of narrative change essential to comprehending the holistic storyline of videos.
However, direct detection of abrupt changes in the query-frame similarity signal is often unreliable due to high-frequency noise arising from model uncertainty and transient visual variations.
To address this, we leverage the wavelet transform, which provides an ideal solution through its multi-resolution analysis in both time and frequency domains.
By applying this transform, we decompose the noisy signal into multiple scales and extract a clean semantic change signal from the coarsest scale. We identify the local extrema of this signal as semantic boundaries, which segment the video into coherent clips.
Building on this, WFS-SB comprises a two-stage strategy: first, adaptively allocating a frame budget to each clip based on a composite importance score; and second, within each clip, employing the Maximal Marginal Relevance approach to select a diverse yet relevant set of frames.
Extensive experiments show that WFS-SB significantly boosts LVLM performance, e.g., improving accuracy by \textbf{5.5\% on VideoMME, 9.5\% on MLVU, and 6.2\% on LongVideoBench}, consistently outperforming state-of-the-art methods.
Our code is available at \url{https://github.com/MAC-AutoML/WFS-SB}.
\end{abstract}
\section{Introduction}
\label{sec:intro}

\begin{figure}[t]
\centering
\includegraphics[width=\linewidth]{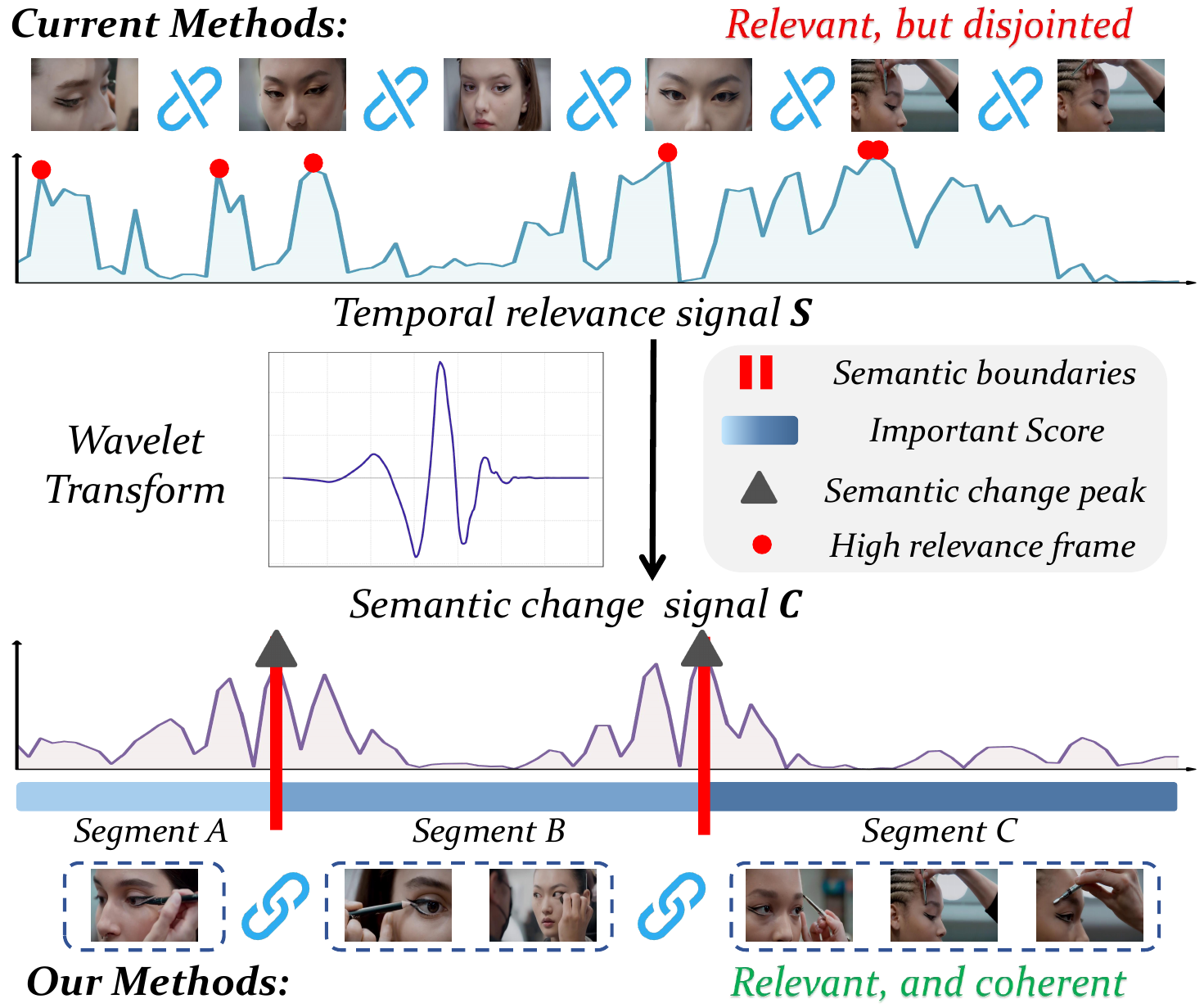}
\caption{
A comparison of frame selection strategies, illustrated with the query: ``What is the process of applying makeup around the eyes in the video?".
(Top) Current approach: Selects scattered, high-relevance frames (e.g., any frame with eyes), which fails to preserve the procedural order.
(Bottom) Our approach: Employs a wavelet-based method to first segment the video into semantically coherent segments (e.g., applying eyeliner, shaping eyebrows) and then samples frames from each segment. This preserves the semantic integrity essential for process comprehension.
}
\label{fig:task}
\end{figure}

The remarkable success of Large Language Models (LLMs)~\cite{touvron2023llama, achiam2023gpt4} has catalyzed the development of powerful Large Vision-Language Models (LVLMs) capable of sophisticated multimodal reasoning~\cite{liu2023visual, zhu2023minigpt, wang2024qwen2vl}, with recent extensions into the video domain~\cite{maaz2024video, lin2024video}. While demonstrating impressive performance on short clips, applying these models to long-form videos confronts a fundamental bottleneck: the massive number of frames clashes with limited computational budgets and fixed-size context windows. 

Directly processing every frame is infeasible, necessitating a strategy to condense the visual stream. Consequently, identifying a concise subset of representative frames, known as keyframe selection, has become a critical prerequisite for deploying LVLMs on real-world, long-form videos. While alternatives like extending context windows~\cite{liu2024world,zhang2024long} or video-to-text summarization~\cite{Park26LVNet, ma2025drvideo, fan2024videoagent} exist, they often incur prohibitive computational overhead or risk significant information loss. Frame selection, therefore, presents a more direct and balanced approach, but it begs the question: what constitutes an optimal selection?


Existing methods typically select frames with the highest relevance scores to a given query. This approach, however, disregards the narrative structure of the video, akin to understanding a book's plot by selecting its most beautifully illustrated pages. The resulting frame collection, though individually relevant, is disjointed and fails to capture causality, development, or process. For instance, as illustrated in Figure~\ref{fig:task}, existing methods often retrieve scattered frames showing eyes but fail to reflect the sequential procedure. In this paper, we introduce \textbf{W}avelet-based \textbf{F}rame \textbf{S}election by Detecting \textbf{S}emantic \textbf{B}oundary (\textbf{WFS-SB}), a training-free framework that presents a new perspective: effective video understanding should not only focus on ``which frames have high relevance," but rather on ``when do the story's chapters change?". These chapter change moments manifest as abrupt shifts in the query-frame similarity signal (e.g., transitioning from ``cooking" to ``eating"). However, this raw signal is often fraught with high-frequency noise from model uncertainty or incidental visual variations, making the direct detection of these boundaries unreliable.

Confronted with this challenge, we reframe boundary detection as a signal processing problem. The wavelet transform emerges as an ideal tool for analyzing this non-stationary and noisy relevance signal. Its capacity for multi-resolution analysis in both time and frequency domains allows us to effectively distinguish meaningful, macro-level semantic shifts from spurious, high-frequency noise. By applying a multi-level wavelet decomposition, we extract a clean semantic change signal from the coarsest scale, whose local extrema precisely identify the temporal semantic boundaries. These boundaries effectively segment the video into coherent sections, akin to chapters in a story.


Based on this segmentation, WFS-SB implements a two-stage selection strategy: first, adaptively allocating a frame budget to each clip based on a composite importance score; and second, within each clip, employing the Maximal Marginal Relevance approach to select a diverse yet relevant set of frames. 
Extensive experiments demonstrate that WFS-SB significantly boosts LVLM performance across multiple benchmarks. For instance, when applied to LLaVA-Video-7B with 8 input frames, it improves accuracy by \textbf{5.5\% on VideoMME, 9.5\% on MLVU, and 6.2\% on LongVideoBench}, consistently outperforming state-of-the-art methods.

Our contributions are threefold:
\begin{itemize}[leftmargin=*, noitemsep, topsep=0pt]
    \item We introduce a new perspective for frame selection that prioritizes capturing semantic shifts over merely high-relevance. We formulate it as a signal processing problem and employ wavelet transform to robustly detect semantic boundaries from noisy relevance signals.
    \item We propose WFS-SB, a training-free framework that advances frame selection from isolated relevance sampling to a semantic structure-aware process, incorporating semantic segmentation, adaptive budget allocation, and intra-clip diversification.
    \item We conduct extensive experiments demonstrating that WFS-SB achieves state-of-the-art performance on multiple long-video benchmarks, validating the superiority of our semantic-shift-first approach.
\end{itemize}

\section{Related Work}
\label{sec:related}


\noindent\textbf{Efficient Architectures and Strategies for Long Video Understanding.}
The evolution from early CNN-RNN models to contemporary Large Vision-Language Models (LVLMs)~\cite{liu2023visual, zhu2023minigpt,bai2025qwen2.5vl} reflects a persistent drive to process increasingly complex visual data. To specifically address long videos, researchers have pursued two primary avenues. The first centers on architectural innovations, such as hierarchical attention or memory mechanisms that expand model context windows~\cite{liu2024world,zhang2024long, shu2025video}. The second involves strategic abstraction, including converting video into text summaries~\cite{wang2025videotree, ma2025drvideo,fan2024videoagent,luo2024video} or pruning and merging visual tokens~\cite{shen2024longvu,luo2025quota,li2024llama,song2026ktv}. While powerful, these approaches often introduce significant computational overhead or risk discarding critical visual details during abstraction. This positions lightweight pre-processing, particularly frame selection, as a vital and pragmatic complementary strategy.


\noindent\textbf{Applications of the Wavelet Transform.}
The wavelet transform's capacity for localized time-frequency analysis has made it indispensable across numerous disciplines. It is a cornerstone of signal and image processing for tasks like denoising and compression, as seen in the JPEG 2000 standard~\cite{zhang2019wavelet,grobbelaar2022survey, tian2023multi}. Its multi-scale nature is also well-suited for analyzing complex physical phenomena, such as turbulence and seismic signals~\cite{farge1992wavelet}. Recently, wavelets have been integrated into machine learning, serving as powerful feature extractors and enhancing deep learning architectures~\cite{sifuzzaman2009application}. In this work, we creatively adapt the wavelet transform to a new domain: identifying semantic boundaries within a query-relevance signal for video frame selection.


\noindent\textbf{Frame Selection for Long Video Understanding.}
Frame selection methods offer a practical solution to the constraints of fixed context windows in LVLMs. While early techniques relied on low-level, query-agnostic cues like shot boundaries~\cite{nasreen2013key,sheena2015key}, modern approaches are more sophisticated. We broadly divide them into training-based and training-free methods. Training-based approaches learn a selection policy but can be data-hungry and computationally expensive~\cite{buch2025flexible, yu2025framevoyager,li2025frameoracle}. Consequently, training-free methods have gained traction for their flexibility. Prominent strategies include inverse transform sampling (BOLT~\cite{liu2025bolt}), Markov decision processes (MDP3~\cite{sun2025mdp3}), joint relevance-coverage optimization (AKS~\cite{tang2025adaptive}), subgraph selection (KFC~\cite{fang2025threading}), dynamic resolution processing (Q-frame~\cite{zhang2025qframe}, VQOS~\cite{wutraining}), temporal object-centric search (T*~\cite{ye2025re}), and iterative reasoning (A.I.R.~\cite{zou2025air}). However, most existing methods still focus on sampling individually relevant frames, often failing to capture the video's overarching semantic structure. Our work, WFS-SB, addresses this gap by first employing a wavelet-driven, training-free approach to detect semantic boundaries, thereby preserving the video's structural integrity before selection.

\section{Method}
\label{sec:method}

\begin{figure*}[t]
\centering
\includegraphics[width=\linewidth]{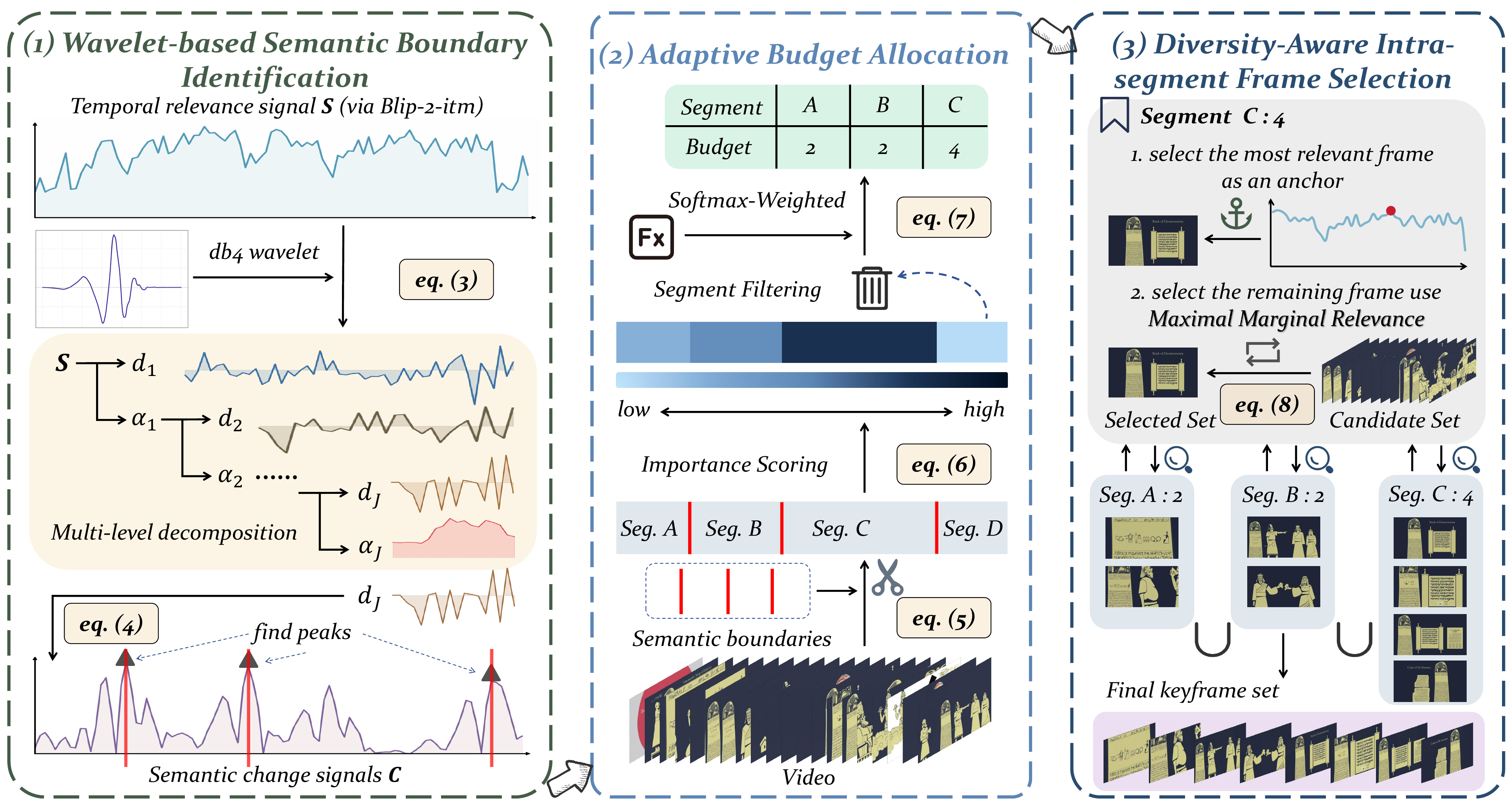}
\caption{
An overview of our proposed WFS-SB framework. The process unfolds in three main stages: (1) Wavelet-based Semantic Boundary Identification: The raw query-frame relevance signal is decomposed using a multi-level Discrete Wavelet Transform. We isolate and reconstruct the coarsest detail coefficients to generate a robust semantic change signal, whose peaks define the boundaries of coherent semantic segments. (2) Adaptive Budget Allocation: A composite importance score is computed for each segment, which guides a softmax-weighted distribution of the total frame budget $K$ = 8. (3) Diversity-Aware Intra-segment Selection: Finally, a localized Maximal Marginal Relevance (MMR) selection is performed within each segment to ensure both relevance and diversity in the final keyframe set.
}
\label{fig:pipeline}
\end{figure*}
\subsection{Problem Formulation}
\label{sec:formulation}

Given a video $\mathcal{V}$ consisting of $T$ frames and a user query $q$, the task of frame selection is to extract a concise yet representative frame subset $\mathcal{F} = \{f_1, f_2, \dots, f_K\}$, where the indices are chronologically ordered and $K \ll T$. The value of $K$ is generally determined in advance, considering both the context length constraints of large vision-language models (LVLMs) and the specific demands of downstream applications or user preferences. The ultimate objective is to choose $\mathcal{F}$ such that, when it serves as visual context together with the query $q$, the LVLM can produce the most accurate and informative response possible.

To quantify the semantic relevance of each frame to the query, we first uniformly subsample the video at 1 frame-per-second (fps), yielding a sequence of $N$ frames. For each sampled frame $f_t$, we compute an Image-Text Matching (ITM) score using a pretrained vision-language model:
\begin{equation}
s_t = \mathcal{M}(q, f_t), \quad t = 1, \dots, N,
\label{eq:itm_score}
\end{equation}
where $\mathcal{M}$ denotes the ITM head of BLIP-2~\cite{li2023blip2}. Notably, we only use its ITM head for score computation, without involving the large-scale decoder part. These scores form the foundation for our subsequent analysis.
\subsection{Modeling Temporal Relevance Signal}
\label{sec:signal_modeling}

Rather than treating the ITM scores $\{s_t\}_{t=1}^N$ as isolated relevance measurements, we adopt a signal processing perspective: these scores collectively form a ``temporal relevance signal"—a continuous waveform that encodes how the video's semantic alignment with the query evolves over time. This reframing enables us to leverage powerful tools from signal analysis to identify structural transitions in video content.
However, this temporal relevance signal exhibits three challenging characteristics:

\textbf{Non-stationarity.} This is the most critical feature of the signal. The dynamic nature of video content dictates that the signal's statistical properties change drastically over time. A semantic segment highly relevant to the query (e.g., ``a person hugging another") might occur within a few seconds, causing a sharp local fluctuation in the signal. In contrast, during irrelevant segments, the signal may remain stable at a low level. Traditional global analysis tools like the Fourier transform are unsuitable as they cannot capture the temporal location of these changes.

\textbf{Multi-scale structure.} Semantically coherent segments in a video naturally possess different temporal scales. A rapid action (e.g., ``waving a hand") might span only a few frames, manifesting as a sharp, transient pulse in the signal. Conversely, a prolonged process (e.g., ``the entire cooking procedure") could extend over hundreds of frames, appearing as a slow, broad peak. We require an analytical tool that can resolve both macroscopic trends and microscopic details simultaneously.

\textbf{Low signal-to-noise ratio.} The raw relevance signal is inundated with substantial noise. This noise originates from several sources: 1) Model uncertainty: Minor perturbations in the internal representations of a VLM for visually similar yet semantically indistinct frames can lead to score jitter. 2) Cross-modal ambiguity: The semantic alignment between a textual query and visual content is inherently ambiguous. 3) Visual artifacts: Factors unrelated to the query's semantics, such as changes in lighting, object occlusions, or camera motion, can cause unintended score fluctuations. This high-frequency noise severely interferes with the identification of true semantic boundaries, making methods based on simple amplitude thresholds or naive gradient detection highly prone to failure.

These three properties collectively motivate our choice of wavelet transform. Wavelets provide a mathematically principled framework that simultaneously addresses all three challenges through multi-resolution decomposition in both the time and frequency domains.

\subsection{Wavelet-based Frame Selection}
\label{sec:bfswt_framework}

As illustrated in Figure~\ref{fig:pipeline}, the method integrates three core components in a hierarchical pipeline: 1) Wavelet-Based Semantic Boundary identification—partitioning the video into temporally coherent segments by identifying semantic boundaries through wavelet transform. 2) Adaptive Budget Allocation—computing segment importance scores and distributing frame budget non-uniformly across segments. 3) Diversity-Aware Intra-segment Frame Selection—applying diversity-driven selection within each segment's allocated budget. This hierarchical design preserves narrative structure and ensures local detail and global coherence.
\subsubsection{Wavelet-based Semantic Boundary Identification}
\label{sec:wavelet_decomposition}
To address the challenges outlined in Section~\ref{sec:signal_modeling}, we employ the Discrete Wavelet Transform (DWT). Its multi-resolution analysis is uniquely suited to decompose the similarity score sequence, allowing us to isolate large-scale semantic transitions from transient noise and accurately identify semantic boundaries.

\noindent\textbf{Adaptive Multi-level Decomposition.}
We apply the DWT to the ITM score sequence $\{s_t\}_{t=1}^N$ using the Daubechies-4 (db4) wavelet, which offers a favorable trade-off between smoothness and compact support. The DWT recursively filters the signal, splitting it at each level $j$ into low-frequency approximation coefficients $a_j$ and high-frequency detail coefficients $d_j$. The key parameter is the decomposition level $J$, which we set adaptively based on the sequence length $N$:
\begin{equation}
J = \max\left(1, \left\lfloor \log_2 N \right\rfloor - l\right),
\label{eq:adaptive_level}
\end{equation}
where $l$ is a drift factor controlling the decomposition depth, set to 3. This adaptive strategy ensures the analysis scale is appropriate for the signal's duration; it uses a larger $J$ for longer videos to focus on coarse, stable components and a smaller $J$ for shorter videos to retain crucial temporal details. The final DWT output is a multi-scale representation of the original signal:
\begin{equation}
\text{DWT}(s_t) = \{a_J, d_J, d_{J-1}, \dots, d_1\}.
\label{eq:dwt}
\end{equation}

\noindent\textbf{Semantic Change Feature Extraction.}
Semantic boundaries are characterized by significant, large-scale shifts in the ITM score, rather than by high-frequency fluctuations. These slow-varying but substantial changes are captured by the detail coefficients at the coarsest scale, $d_J$. To isolate this specific feature, we perform a reconstruction using the Inverse Discrete Wavelet Transform (IDWT), where only the $d_J$ coefficients are retained and all others are zeroed out:
\begin{equation}
\tilde{s}_t = \text{IDWT}(\{\mathbf{0}, d_J, \mathbf{0}, \dots, \mathbf{0}\}).
\label{eq:wavelet_reconstruction}
\end{equation}
The resulting signal, $\{\tilde{s}_t\}$, represents the component of the ITM score that reflects coarse-grained temporal variations, effectively acting as a robust indicator of semantic change while naturally suppressing high-frequency noise concentrated in finer-scale coefficients (e.g., $d_1, d_2$).

\noindent\textbf{Semantic Boundary Detection.}
The moments of most significant semantic change correspond to local extrema in the magnitude of the change indicator signal, $\tilde{s}_t$. We therefore define a change intensity signal as $c_t = |\tilde{s}_t|$ and apply a peak detection algorithm to it. This process identifies the temporal indices of significant boundaries while filtering out spurious peaks from residual noise using adaptive height and prominence thresholds. Let $\mathcal{B} = \{b_1, \dots, b_M\}$ be the set of detected boundary indices. These boundaries partition the video into $M+1$ Temporal Semantic Segments:
\begin{equation}
\mathcal{G} = \{[1, b_1], [b_1+1, b_2], \dots, [b_M+1, N]\},
\label{eq:segments}
\end{equation}
where each segment represents a temporally coherent unit with stable relevance to the query. These segments form the structural basis for subsequent frame selection.
\subsubsection{Adaptive Budget Allocation}
\label{sec:budget_allocation}

Once the temporal semantic segments are established, we allocate the total frame budget $K$ according to the semantic importance of each segment. This adaptive strategy ensures that semantically critical segments receive dense coverage, optimizing the trade-off between global video understanding and fine-grained local detail.

\noindent\textbf{Segment Importance Score.}
We compute a composite importance score, $\text{Imp}(\mathcal{G}_i)$, for each temporal semantic segment $\mathcal{G}_i$ by integrating multiple relevance cues:
\begin{equation}
\text{Imp}(\mathcal{G}_i) = w_d \cdot \frac{|\mathcal{G}_i|}{N} + w_a \cdot \bar{s}_i + w_m \cdot s_i^{\max} + w_v \cdot \frac{\sigma_i^2}{\sigma_{\text{global}}^2},
\label{eq:importance}
\end{equation}
where $|\mathcal{G}_i|/N$ represents the normalized segment duration, $\bar{s}_i$ is the mean relevance score, $s_i^{\max}$ is the max relevance score, and $\sigma_i^2 / \sigma_{\text{global}}^2$ is the ratio of the relevance score variance within the segment to the global relevance score variance. These terms collectively measure a segment's duration, average importance, peak significance, and internal content diversity. The coefficients $w_d, w_a, w_m, w_v$ are weighting factors, set to $0.4, 0.2, 0.3, 0.1$ by default.

\noindent\textbf{Segment Filtering.}
To concentrate the budget on salient content, we prune segments with low importance scores. Specifically, we discard any segment $\mathcal{G}_i$ for which $\text{Imp}(\mathcal{G}_i)$ falls below an adaptive threshold $\tau$, defined as $\tau$ = $\text{mean}({\text{Imp}})$ - $\eta \cdot \text{std}(\text{Imp})$. The parameter $\eta$ controls the filtering aggressiveness and is set to 1.2.

\noindent\textbf{Softmax-Weighted Budget Distribution.}
We then employ a softmax function over the importance scores to assign a proportional budget, $k_i$, to each remaining segment $\mathcal{G}_i$:
\begin{equation}
k_i = \left\lfloor K \cdot \frac{\exp(\text{Imp}(\mathcal{G}_i))}{\sum_{j} \exp(\text{Imp}(\mathcal{G}_j))} \right\rfloor,
\label{eq:budget_allocation}
\end{equation}
To ensure the allocation is exact ($\sum k_i$ = $K$), any frames remaining from the floor operation are assigned greedily to the segments with the largest fractional parts.
\subsubsection{Diversity-Aware Intra-segment Frame Selection}
\label{sec:frame_selection}
With a budget $k_i$ allocated to each temporal semantic segment $\mathcal{G}_i$, the final step is to select the specific frames from within that segment. This localized selection strategy is critical: it prevents frames from one segment being suppressed by visually similar frames from another segment, while maintaining frame diversity within each segment.

For each segment $\mathcal{G}_i$, we first select its most relevant frame as an anchor. We identify the index of this frame, $t_{\text{anchor}}$, and initialize the set of selected indices for the segment, $\mathcal{T}_i$, with it:
$t_{\text{anchor}}$ = $\argmax_{t \in \mathcal{G}_i} s_t$, $\mathcal{T}_i = \{t_{\text{anchor}}\}$.
To select the remaining $k_i - 1$ frames, we employ a localized Maximal Marginal Relevance (MMR) approach~\cite{carbonell1998use}. This method iteratively identifies the next best frame index, $t^*$, by balancing relevance and diversity:
\begin{equation}
t^* = \underset{t \in \mathcal{G}_i \setminus \mathcal{T}_i}{\operatorname{argmax}} \left[ \lambda \cdot s_t - (1-\lambda) \cdot \max_{t' \in \mathcal{T}_i} \text{sim}(f_t, f_{t'}) \right],
\label{eq:mmr}
\end{equation}
where $\text{sim}(f_t, f_{t'})$ is the cosine similarity between the visual embeddings of candidate frame $f_t$ and a previously selected frame $f_{t'}$. The parameter $\lambda$ (set to 0.5) balances the trade-off between relevance and diversity. After finding the optimal index $t^*$, it is added to the set $\mathcal{T}_i$. This process is repeated until $|\mathcal{T}_i| = k_i$. The final selected keyframe subset for the video is the union of frames corresponding to all selected indices across all segments.
\section{Experiments}
\label{sec:experiments}


\begin{table*}[t]
\centering
\small  
\caption{Comparison of VLMs and various frame selection methods on VideoMME, MLVU, and LVB. We report accuracy scores (\%). \textbf{Bold} indicates best performance and \underline{underline} indicates second-best performance in each model-dataset group. $^{\dagger}$Reproduced results, $^{*}$Reported results, $^{\ddagger}$Training-based methods, $^{\S}$LVLM-dependent iterative method.}
\label{tab:main_results}
\setlength{\tabcolsep}{3pt}  
\begin{tabular}{l|l|c|c|ccc|ccc|ccc}
\toprule
\multirow{2}{*}{\textbf{Model}} & \multirow{2}{*}{\textbf{Method}} & \multirow{2}{*}{\textbf{Size}} & \multirow{2}{*}{\textbf{Frame}} & \multicolumn{3}{c|}{\textbf{VideoMME}~\cite{fu2025videomme}} & \multicolumn{3}{c|}{\textbf{MLVU}~\cite{zhou2024mlvu}} & \multicolumn{3}{c}{\textbf{LongVideoBench}~\cite{wu2024longvideobench}} \\
\cmidrule(lr){5-7} \cmidrule(lr){8-10} \cmidrule(lr){11-13}
& & & & Base & +Method & $\Delta$ & Base & +Method & $\Delta$ & Base & +Method & $\Delta$ \\

\midrule

\multirow{6}{*}{\makecell[l]{LLaVA-OV~\cite{li2024llavaov}}}
& Frame-Voyager$^{*\ddagger}$~\cite{yu2025framevoyager} & 7B & 8 & 53.3 & 57.5 & \gain{+4.2}  & 58.5 & 65.6 & \gain{+7.1} & - & - & - \\
& KFC$^{*}$~\cite{fang2025threading} & 7B & 8 & 53.3 & 55.4 & \gain{+2.1} & 58.5 & \underline{66.2} & \gain{\underline{+7.7}} & 54.5 & 55.6 & \gain{+1.1} \\
& BOLT$^{*}$~\cite{liu2025bolt} & 7B & 8 & 53.8 & 56.1 & \gain{+2.3} & 58.9 & 63.4 & \gain{+4.5} & 54.2 & 55.6 & \gain{+1.4} \\
& AKS$^{\dagger}$~\cite{tang2025adaptive} & 7B & 8 & 54.1& \underline{58.2} & \gain{\underline{+4.1}} & 58.6 & 62.9 & \gain{+4.3} & 54.2 & \underline{58.4} & \gain{\underline{+4.2}} \\
& FrameOracle$^{*\ddagger}$~\cite{li2025frameoracle}& 7B & 8 & 53.8 & 57.5 & \gain{+3.7}  & 58.4 & 62.9 & \gain{+4.5} & 54.3 & 56.0 & \gain{+1.7} \\
& \hl WFS-SB$^{\dagger}$ & \hl 7B & \hl 8 & \hl 54.1 & \hl\textbf{59.3} & \hl\gain{\textbf{+5.2}} & \hl 58.6 & \hl\textbf{67.2} & \hl\gain{\textbf{+8.6}} & \hl 54.2 & \hl\textbf{59.8} & \hl\gain{\textbf{+5.6}} \\

\midrule

\multirow{5}{*}{LLaVA-Video~\cite{zhang2024llavavideo}}
& KFC$^{*}$ & 7B & 8 & 55.9 & 57.6 & \gain{+1.7} & 60.5 & \textbf{66.9} & \gain{+6.4} & 54.2 & 56.5 & \gain{+2.3} \\
& BOLT$^{*}$ & 7B & 8 & 56.0 & 58.6 & \gain{+2.6} & - & - & - & - & - & - \\
& AKS$^{\dagger}$ & 7B & 8 & 56.2& \underline{60.1} & \gain{\underline{+3.9}} & 57.4& 64.2 & \gain{\underline{+6.8}} & 54.9 & \underline{59.6} & \gain{\underline{+4.7}}  \\
& FrameOracle$^{*\ddagger}$ & 7B & 8 & 55.9&58.9 & \gain{+3.0} & 60.5 & 63.4 & \gain{+2.9} & 54.2 &56.9 & \gain{+2.7} \\
& \hl WFS-SB$^{\dagger}$ & \hl 7B & \hl 8 & \hl 56.2 & \hl\textbf{61.7} & \hl\gain{\textbf{+5.5}} & \hl 57.4 & \hl\textbf{66.9} & \hl\gain{\textbf{+9.5}} & \hl 54.9 & \hl\textbf{61.1} & \hl\gain{\textbf{+6.2}} \\

\midrule

\multirow{4}{*}{Qwen2.5-VL~\cite{bai2025qwen2.5vl}}
& AKS$^{\dagger}$ & 7B & 32 & 61.2& 64.0 & \gain{+2.8} & 59.7 & 67.2 & \gain{+7.5} & 58.9 & \underline{63.2} & \gain{\underline{+4.3}} \\
& MDP$^3$\textsuperscript{*}~\cite{sun2025mdp3, zou2025air} & 7B & 32 & 60.8 & 63.8 & \gain{+3.0} & 59.3 & 66.2 & \gain{+6.9} & 58.1 & 60.0 & \gain{+1.9}\\
& A.I.R.$^*$~\cite{zou2025air} & 7B & $\leq$32$^{\S}$ &  60.8& \textbf{65.0} & \gain{\textbf{+4.2}} & 59.3 & \underline{67.5} & \gain{\underline{+8.2}} & 58.1 & 61.4 & \gain{+3.3}\\
& \hl WFS-SB$^{\dagger}$ & \hl 7B & \hl 32 & \hl 61.2 & \hl\underline{64.4} & \hl\gain{\underline{+3.2}} & \hl 59.7 & \hl\textbf{70.4} & \hl\gain{\textbf{+10.7}} & \hl 58.9 & \hl\textbf{64.4} & \hl\gain{\textbf{+5.5}} \\

\midrule

\multirow{4}{*}{InternVL-3~\cite{zhu2025internvl3}}
& AKS$^{\dagger}$ & 8B & 32 & 65.6& 66.3 & \gain{+0.7} & \textbf{68.4} & 74.2 & \gain{+5.8} & 58.5 & 61.5 & \gain{+3.0} \\
& MDP$^3$\textsuperscript{*} & 8B & 32 & 65.6& 66.8 & \gain{+1.2} & 68.4 & 74.0 & \gain{+5.6} & 58.3 & 60.9 & \gain{+2.6} \\
& A.I.R.$^*$ & 8B & $\leq$32$^{\S}$ & 65.6& \textbf{68.2} & \gain{\textbf{+2.6}} & 68.4 & \underline{74.5} & \gain{\underline{+6.1}} & 58.3 & \underline{62.8} & \gain{\textbf{+4.5}} \\
& \hl WFS-SB$^{\dagger}$ & \hl 8B & \hl 32 & \hl 65.6 & \hl\underline{67.4} & \hl\gain{\underline{+1.8}} & \hl 68.4 & \hl\textbf{74.8} & \hl\gain{\textbf{+6.4}} & \hl 58.5 & \hl\textbf{62.9} & \hl\gain{\underline{+4.4}} \\

\bottomrule
\end{tabular}
\end{table*}


\subsection{Experimental Setup}



\noindent\textbf{Benchmarks and Backbone Models.} We evaluate WFS-SB on three long-video QA benchmarks: VideoMME~\cite{fu2025videomme} (900 videos, avg. 17 min, 2,700 QA pairs), MLVU~\cite{zhou2024mlvu} (2,174 questions across 7 categories, avg. 11 min), and LongVideoBench (LVB)~\cite{wu2024longvideobench} (1,337-pair validation set, avg. 12 min). Subtitles are not used. To demonstrate generalizability, we integrate WFS-SB with four state-of-the-art LVLMs: LLaVA-OneVision-7B~\cite{li2024llavaov}, LLaVA-Video-7B~\cite{zhang2024llavavideo}, Qwen2.5-VL-7B~\cite{bai2025qwen2.5vl}, and InternVL3-8B~\cite{zhu2025internvl3}, spanning diverse architectures and training philosophies.




\noindent\textbf{Baselines.} We benchmark WFS-SB against a suite of state-of-the-art frame selection methods, including Frame-Voyager~\cite{yu2025framevoyager}, KFC~\cite{fang2025threading}, BOLT~\cite{liu2025bolt}, AKS~\cite{tang2025adaptive}, FrameOracle~\cite{li2025frameoracle}, MDP$^3$~\cite{sun2025mdp3}, and A.I.R.~\cite{zou2025air}. For methods with publicly available code ($\dagger$), we re-evaluate them in our controlled environment to ensure a direct comparison. For others ($*$), we cite the results reported in their original publications under comparable settings.




\noindent\textbf{Implementation Details.} Candidate frames are uniformly sampled from each video at 1 frame per second (1 FPS). We use BLIP-2-ITM-ViT-g~\cite{li2023blip2} to compute query-to-frame matching scores. The wavelet decomposition drift factor $l$ is set as 3 with Daubechies-4 (db4) wavelet. Budget allocation weights are $w_d$=0.4, $w_a$=0.2, $w_m$=0.3, $w_v$=0.1, The filtering aggressiveness factor $\eta$ is set to 1.2, and the MMR parameter is $\lambda=0.5$. All experiments are conducted with frame budgets $K \in \{8, 16, 32, 64\}$ on NVIDIA A800 GPUs with 80 GB of memory. We report accuracy measured by the LMMs-Eval toolkit~\cite{zhang2024lmmseval}.



\subsection{Comparison with State-of-the-Art}


Table~\ref{tab:main_results} presents comprehensive results across all benchmarks, LVLM architectures, and frame budgets. WFS-SB demonstrates \textbf{powerful plug-and-play enhancement}, delivering consistent improvements without requiring model fine-tuning or architectural modifications—average gains of 3.9\% on VideoMME, 8.8\% on MLVU, and +5.4\% on LongVideoBench across all configurations validate its universal applicability. Our method achieves \textbf{state-of-the-art accuracy with superior efficiency}, outperforming both training-based methods (Frame-Voyager, FrameOracle) and iterative approaches (A.I.R.) while maintaining training-free deployment and single-pass inference. Most notably, WFS-SB exhibits \textbf{robust performance across diverse benchmarks}, with particularly strong gains under tight frame budgets (e.g., for LLaVA-Video-7B at $K$=8, gains are 9.5\% on MLVU, 6.2\% on LongVideoBench and 5.5\% on VideoMME), demonstrating that our wavelet-based frame selection effectively adapts to varying content complexity and temporal scales.



\subsection{Ablation Study}



\begin{figure}[t]
\centering
\includegraphics[width=\linewidth]{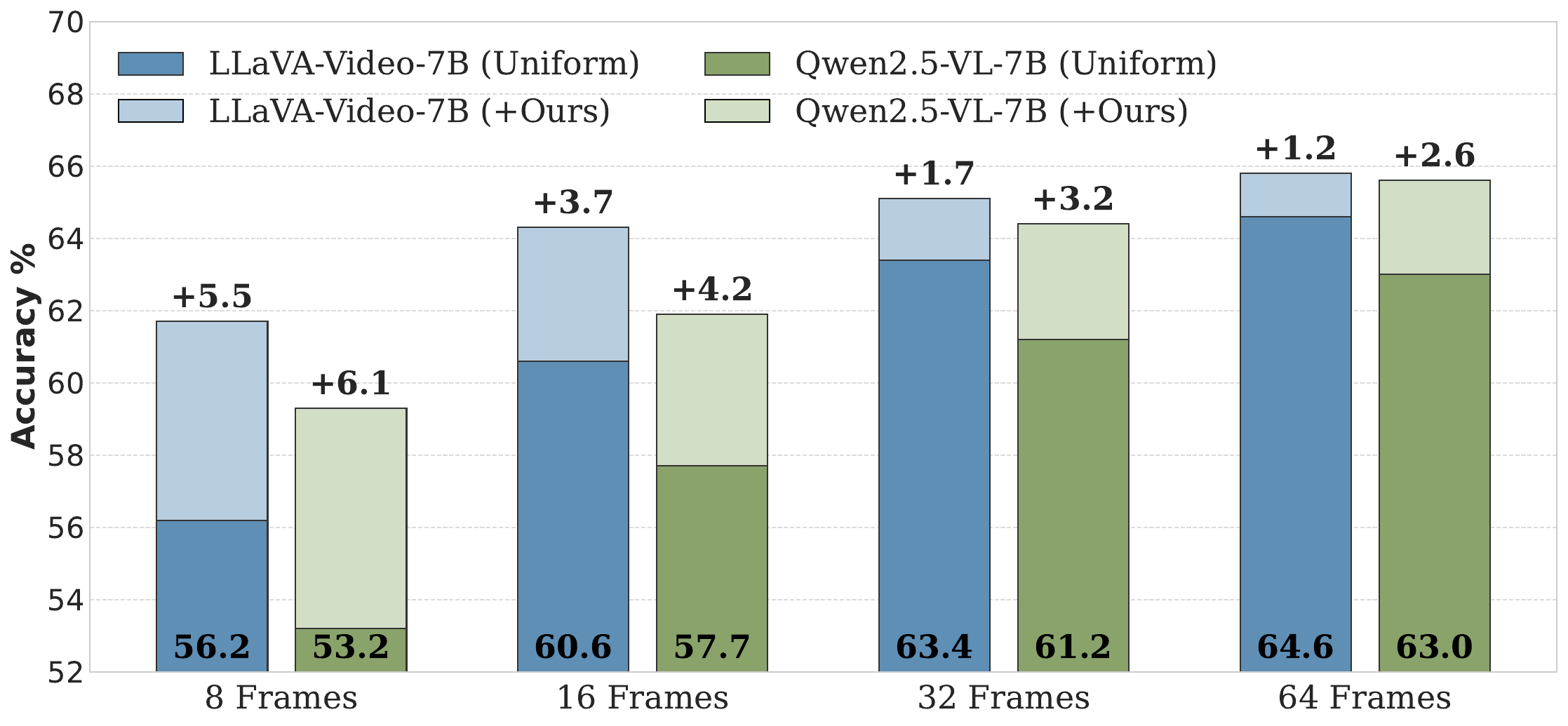}
\caption{Performance comparison across different frame budgets K on VideoMME. WFS-SB consistently outperforms uniform sampling across all LVLMs and budget settings.}
\label{fig:frame_budget}
\vspace{-1mm}
\end{figure}

\noindent\textbf{Different Sampling Frames.}
Figure~\ref{fig:frame_budget} illustrates the performance of WFS-SB across frame budgets \{8, 16, 32, 64\} on VideoMME with multiple backbones. We observe consistent improvements over uniform sampling (+1.2\% to +5.5\%), with particularly pronounced gains at smaller budgets ($K$=8, 16). This validates that boundary-aware selection becomes increasingly critical when frame resources are constrained, demonstrating strong generalization across different VLMs and budget sizes.

\noindent\textbf{Different LVLM Scales.}
To evaluate scalability, we test WFS-SB on Qwen2.5-VL at four scales (3B, 7B, 32B, 72B). Table~\ref{tab:model_scale} shows consistent improvements with average gains of 3.3\%, 3.7\%, +3.0\%, and 2.8\% respectively. The consistent gains validate that WFS-SB's benefits are orthogonal to model scale, making it equally valuable for both resource-constrained deployments (smaller models) and performance-critical applications (larger models).

\begin{table}[t]
\centering
\small
\caption{Ablation of different LVLM scales using Qwen2.5-VL on VideoMME. Accuracy (\%) is reported. Results show performance with $K$=16 frames (left) and $K$=32 frames (right). $\Delta$ column shows improvement gains for 16 and 32 frames respectively.}
\label{tab:model_scale}
\begin{tabular}{lccc}
\toprule
\textbf{Model Scale} & \textbf{Baseline} & \textbf{+WFS-SB} & \textbf{$\Delta$} \\
\midrule
Qwen2.5-VL-3B  & 54.1/57.8 & 58.4/60.0 & \gain{+4.3}/\gain{+2.2} \\
Qwen2.5-VL-7B  & 57.7/61.2 & 61.9/64.4 & \gain{+4.2}/\gain{+3.2} \\
Qwen2.5-VL-32B & 60.2/62.9 & 63.0/66.1 & \gain{+2.8}/\gain{+3.2} \\
Qwen2.5-VL-72B & 63.3/66.2 & 66.1/68.9 & \gain{+2.8}/\gain{+2.7} \\
\bottomrule
\end{tabular}
\end{table}
\noindent\textbf{Different VLMs for Query-Frame Similarity.}
We investigate different VLMs for computing query-frame relevance. Table~\ref{tab:vlm_comparison} compares four widely-used VLMs: BLIP-ITM, CLIP-VIT-B, SigLIP-so400m, and BLIP-2-ITM. All VLMs demonstrate consistent performance gains when integrated with WFS-SB (even on CLIP-VIT-B, there is a 3.6\% improvement), validating the robustness and generalization capability of our wavelet-based framework across different feature extractors. While BLIP-ITM achieves the highest performance on VideoMME (62.8\%) and LongVideoBench (63.2\%), we select BLIP-2-ITM as default because it provides more consistent average benefits across diverse settings. The greater success of the BLIP series highlights the benefit of dedicated image-text matching due to its explicit image-text alignment objective.

\begin{table}[t]
\centering
\small
\caption{Ablation of different VLMs for query-frame similarity scoring with Qwen2.5-VL-7B ($K$=16). Accuracy (\%) is reported. Although BLIP-ITM achieves highest performance on VideoMME and LVB, we select BLIP-2-ITM as default because it provides more consistent average benefits across diverse settings.}
\label{tab:vlm_comparison}
\begin{tabular}{lccc}
\toprule
\textbf{VLM Scorer} & \textbf{VideoMME} & \textbf{MLVU} & \textbf{LVB} \\
\midrule
Uniform & 57.7 & 56.2 & 57.0 \\
\midrule
BLIP-ITM~\cite{li2022blip}             & \textbf{62.8} & 67.7 & \textbf{63.2} \\
CLIP-VIT-B~\cite{radford2021learning}              & 61.7 & 66.5 & 60.6 \\
SigLIP-so400m~\cite{zhai2023sigmoid}            &  61.9&  66.4&61.9  \\
\hl BLIP-2-ITM (Ours)~\cite{li2023blip2} & \hl 61.9 & \hl \textbf{67.9} & \hl 62.5 \\
\bottomrule
\end{tabular}
\end{table}

\noindent\textbf{Component Ablation.}
Table~\ref{tab:ablation} systematically evaluates each component's contribution, with emphasis on wavelet-based boundary detection's critical role. Replacing DWT with naive alternatives—local minima or gradient-based methods—causes substantial degradation (61.9$\rightarrow$60.8 and 61.9$\rightarrow$61.2 on VideoMME; 67.9$\rightarrow$64.6 and 67.9$\rightarrow$66.8 on MLVU), with local minima suffering -3.3\% on MLVU. This validates our core hypothesis: wavelet transform's multi-scale decomposition inherently suppresses high-frequency noise in the similarity signal, isolating genuine semantic transitions while filtering spurious fluctuations. In contrast, raw methods suffer from noise amplification, causing false boundary detections and fragmented segmentation. Beyond DWT, removing adaptive budget allocation (-0.5\%) or MMR selection (-1.2\% and -5.2\%) also degrades performance, confirming all components synergistically contribute (+4.2\% on VideoMME, +11.7\% on MLVU).

\begin{table}[t]
\centering
\small
\setlength{\tabcolsep}{3pt}
\caption{Ablation of different components with Qwen2.5-VL-7B ($K$=16). We remove each component to measure its contribution. Accuracy (\%) is reported. B.A. means Budget Allocation.}
\label{tab:ablation}
\begin{tabular}{lcc}
\toprule
\textbf{Configuration} & \textbf{VideoMME} & \textbf{MLVU}\\
\midrule
Uniform Sampling & 57.7 & 56.2\\
\midrule
\hl WFS-SB (Our full method) & \hl \textbf{61.9} & \hl \textbf{67.9} \\
\quad w/o DWT (use raw local minima) & 60.8 & 64.6\\
\quad w/o DWT (use raw gradient) & 61.2 & 66.8\\
\quad w/o Adaptive B.A. (Average B.A.) & 61.6& 67.4\\
\quad w/o MMR (topK selection) & 60.9 &66.7\\
\quad w/o MMR (uniform selection) & 59.2 & 62.7\\
\bottomrule
\end{tabular}
\end{table}

\begin{table}[t]
\centering
\small
\setlength{\tabcolsep}{4pt}
\caption{Ablation of wavelet families and decomposition level $l$ with Qwen2.5-VL-7B ($K$=16). Accuracy (\%) is reported.}
\label{tab:wavelet_ablation}
\begin{tabular}{ccccccc}
\toprule
\textbf{Wavelet} & \hl Db4 & Db4& Db8 & Haar& Sym4& Bior3.3 \\
\boldmath{$l$} & \hl 3 &4 &3 &3 &3 &3 \\
\midrule
\textbf{VideoMME}&\hl \textbf{61.9} & 60.4&\textbf{61.9} &61.3 &\textbf{61.9} & \textbf{61.9}  \\
\textbf{MLVU}& \hl 67.9&\textbf{68.3} &66.4 &68.1 &67.6 &68.2   \\
\bottomrule
\end{tabular}
\end{table}

\noindent\textbf{Wavelet Family and Decomposition Level.}
Table~\ref{tab:wavelet_ablation} demonstrates remarkable robustness across wavelet families and validates our adaptive decomposition level formula (Equation~\ref{eq:adaptive_level}). At $l$=3, multiple families achieve highly consistent performance: Db4, Db8, Sym4, and Bior3.3 all reach 61.9\% on VideoMME, while Haar achieves 61.3\%—only 0.6\% variance despite distinct mathematical properties. This confirms multi-scale signal decomposition, rather than specific basis choice, drives effectiveness. We select Db4 with $l$=3 for best balance across benchmarks (61.9\%/67.9\%). The adaptive level formula proves critical: at $l$=4, VideoMME drops to 60.4\% due to over-smoothing, while MLVU improves to 68.3\% due to different temporal scales. The drift factor in Equation~\ref{eq:adaptive_level} automatically adjusts depth based on video length, ensuring optimal noise suppression across diverse durations.

\begin{table}[t]
\centering
\small
\caption{Ablation of hyperparameters on VideoMME and MLVU with Qwen2.5-VL-7B ($K$=16). Accuracy (\%) is reported.}
\label{tab:hyperparameter_ablation}
\begin{tabular}{cccc}
\toprule
$\boldsymbol{w_d}, \boldsymbol{w_a}, \boldsymbol{w_m}, \boldsymbol{w_v}$ & $\boldsymbol{\lambda}$ & \textbf{VideoMME} & \textbf{MLVU}\\
\midrule
\hl 0.4, 0.2, 0.3, 0.1 &\hl 0.5&\hl  61.9 & \hl \textbf{67.9}\\
0.0, 0.3, 0.3, 0.2 &0.5& 61.6 & 64.8\\
0.5, 0.0, 0.3, 0.2 &0.5& 61.8 & 66.8\\
0.5, 0.3, 0.0, 0.2 &0.5& 62.0 & 66.8\\
0.5, 0.2, 0.3, 0.0 &0.5& \textbf{62.1} & 67.4\\
0.4, 0.2, 0.3, 0.1 &0.3& 61.2 & 67.5\\
0.4, 0.2, 0.3, 0.1 &0.7& 61.6 & 66.5\\
\bottomrule
\end{tabular}
\end{table}

\begin{table}[t]
\centering
\small
\caption{Computational efficiency breakdown on VideoMME with Qwen2.5-VL-7B (avg. N=1040 frames, $K$=32). Times measured on a single NVIDIA A800 GPU and two Intel Xeon Silver 4314 CPUs (16 cores), averaged over 900 videos.}
\label{tab:efficiency}
\begin{tabular}{lcc}
\toprule
\textbf{Component} & \textbf{Time (s)} & \textbf{Complexity} \\
\midrule
ITM Signal Extraction & 19.4 & $O(N)$ \\
DWT and Boundary Detection & $\sim$ 0 & $O(N \log N)$ \\
Budget Allocation & $\sim$ 0 & $O(M)$ \\
MMR Selection & 0.7 & $O(NK^2)$ \\
LVLM Inference & 4.4 & - \\
\bottomrule
\end{tabular}
\end{table}

\noindent\textbf{Hyperparameter Sensitivity.}
A critical advantage of WFS-SB is its robustness despite involving multiple hyperparameters. Table~\ref{tab:hyperparameter_ablation} analyzes sensitivity to the importance score weights and MMR diversity parameter. 
Removing any single component degrades performance (-0.5\% to -3.1\% on MLVU), confirming that all four factors—duration, average relevance, peak relevance, and variance—contribute complementary information for segment importance estimation. The relatively small degradation indicates that while each component is necessary, the overall framework is robust to specific weight values.
\textbf{Most remarkably, we use identical settings across all four LVLM backbones, all three benchmarks, and all frame budgets  without task-specific tuning.} 
The MMR parameter $\lambda$ exhibits particularly stable behavior: performance varies by only 0.4\% on VideoMME across $\lambda \in [0.3, 0.7]$, with optimal balance at $\lambda$=0.5. 
This robustness stems from wavelet decomposition inherently adapting to signal characteristics and budget allocation automatically scaling to segment distributions, enabling deployment with default hyperparameters without expensive grid search.
\subsection{Complexity and Time Analysis}
Table~\ref{tab:efficiency} presents a comprehensive analysis of WFS-SB's computational efficiency. For videos on VideoMME (avg. N=1040 frames), the timing breakdown reveals that ITM signal extraction dominates at 19.4s (79\% of total time)—this represents an optimization opportunity as ITM computation could be accelerated through batching, quantization, or distillation of the BLIP-2-ITM model. The wavelet processing pipeline itself (DWT, boundary detection, budget allocation, MMR) adds only 0.7s overhead. Critically, while ITM extraction introduces computational cost compared to naive uniform sampling, this overhead is entirely justified by the substantial performance gains.



\begin{figure}[t]
\centering
\includegraphics[width=\linewidth]{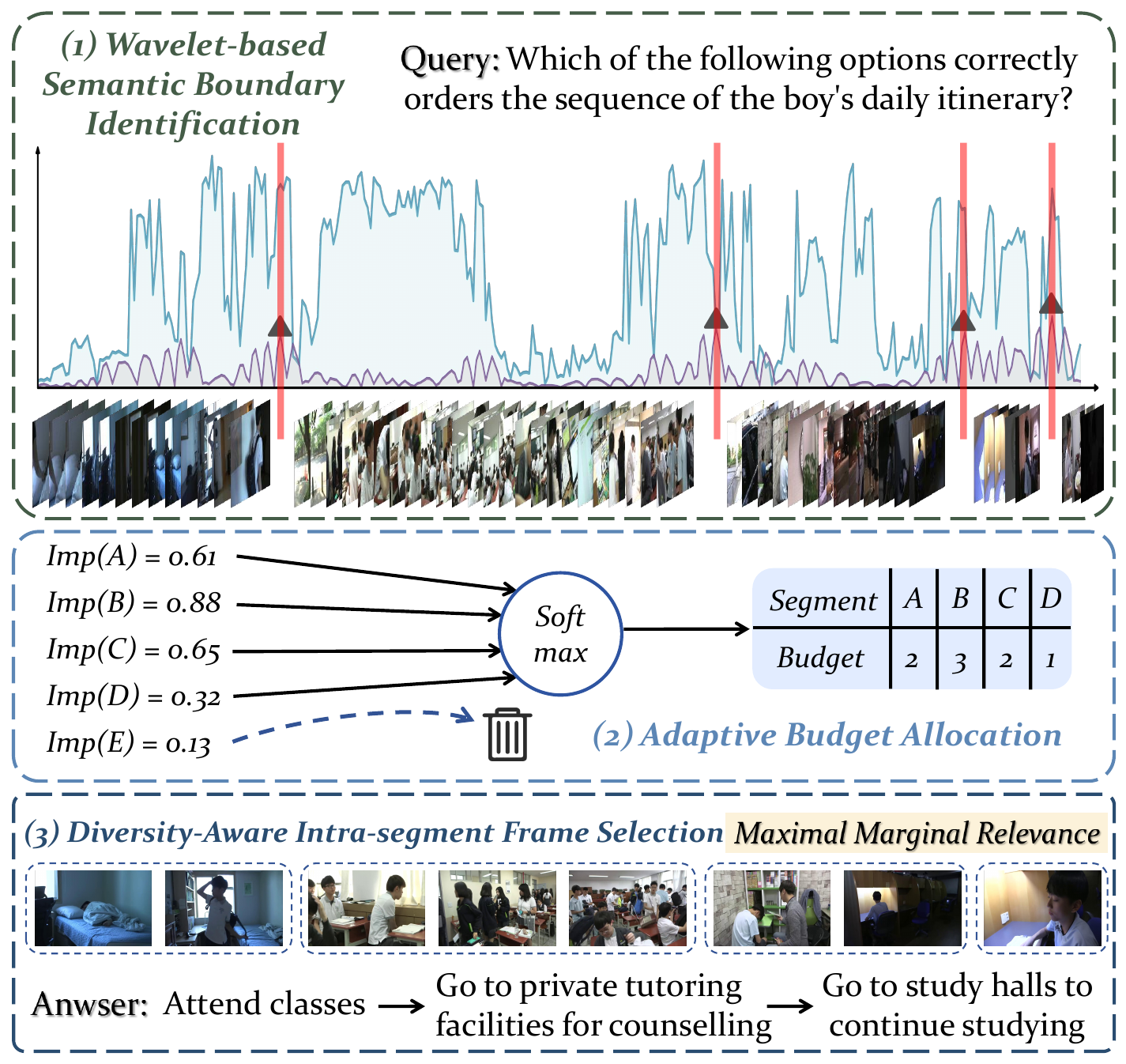}
\caption{Visualization of our WFS-SB framework on a daily itinerary video. Our wavelet-based method first partitions the video into coherent segments by detecting robust semantic boundaries. Subsequently, after filtering segments based on importance scores, Maximal Marginal Relevance sampling is applied within each to select keyframes that preserve the essential narrative, correctly capturing the sequence of class, tutoring, and self-study.}
\label{fig:visualization}
\end{figure}

\subsection{Visualization Analysis}

Figure~\ref{fig:visualization} demonstrates WFS-SB on a daily activities video with distinct yet visually similar events. Our wavelet-based boundary detection identifies key semantic transitions from the noisy similarity signal, segmenting the video into coherent clips. Subsequent stratified sampling ensures selected frames represent each event (class, tutoring, self-study) while preserving chronological narrative, faithfully reconstructing the event sequence and highlighting our method's strength in process-oriented understanding.

\section{Conclusion}
\label{sec:conclusion}



In this paper, we introduced WFS-SB, a training-free framework that enhances long-video understanding for LVLMs by focusing on narrative structure. Our key innovation is the use of wavelet transform to reliably detect semantic boundaries, dividing the video into coherent clips. A two-stage selection strategy then allocates frames based on clip importance and intra-clip diversity. Extensive experiments demonstrate that WFS-SB significantly outperforms state-of-the-art methods across multiple benchmarks, highlighting the importance of capturing a video's narrative flow. 

\section*{Acknowledgments}
This work is supported by the National Key Research and Development Program of China (No.~2025YFE0113500), the National Natural Science Foundation of China (No.~62576299, No.~U21B2037, No.~U22B2051, No.~U23A20383, No.~62176222, No.~62176223, No.~62176226, No.~62072386, No.~62072387, No.~62072389, No.~62002305, No.~62272401, No.~62406071), the Natural Science Foundation of Fujian Province of China (No.~2021J06003, No.~2022J06001), and the Fundamental Research Funds for the Central Universities.
{
    \small
    \bibliographystyle{ieeenat_fullname}
    \bibliography{main}
}

\setcounter{page}{1}
\maketitlesupplementary

\noindent This supplementary material provides extended technical details, additional experiments, and in-depth analyses that complement the main paper. The content is organized as follows:

\noindent\textbf{Section~\ref{sec:suppl_limitations} -- Limitations and Future Work.}
We discuss three main limitations of WFS-SB and outline four promising research directions for extending wavelet-based frame selection.

\noindent\textbf{Section~\ref{sec:further method} -- Further Explanation of the Method.}
We present the detailed mathematical formulation of our adaptive peak detection algorithm.

\noindent\textbf{Section~\ref{sec:suppl_additional_experiments} -- Additional Experimental Results.}
We report extended experimental analyses, including comparisons with additional baseline methods (Section~\ref{sec:suppl_more_baselines}), performance across different frame budgets and models (Section~\ref{sec:suppl_extended_frame_budget}), ablation studies on frame sampling rates (Section~\ref{sec:suppl_fps_ablation}), extended VLM comparisons (Section~\ref{sec:suppl_extended_vlm}), hyperparameter sensitivity analysis (Section~\ref{sec:suppl_segment_filtering}), and qualitative visualizations (Section~\ref{sec:suppl_qualitative}).

\section{Limitations and Future Work}
\label{sec:suppl_limitations}

\subsection{Limitations}
\label{sec:suppl_limitations_current}

While WFS-SB achieves strong performance across multiple benchmarks and architectures, we acknowledge three main limitations.

\noindent\textbf{Computational Overhead of ITM Feature Extraction.}
The primary bottleneck is extracting Image-Text Matching (ITM) scores using BLIP-2. As reported in Table~\ref{tab:efficiency} of the main paper, this accounts for approximately 79\% (19.4 seconds) of preprocessing time. Although the wavelet transform itself is highly efficient ($O(N \log N)$) and boundary detection operates in near-constant time, the upfront cost of computing dense query-frame similarity scores can be prohibitive for extremely long videos or real-time applications. This limitation is shared with other vision-language alignment methods and becomes pronounced for multi-hour videos.

\noindent\textbf{Dependency on Vision-Language Feature Quality.}
Our framework relies on the quality and calibration of ITM scores from a pretrained vision-language model. While Table~\ref{tab:vlm_comparison} demonstrates robustness across multiple VLMs (BLIP-ITM, BLIP-2-ITM, CLIP, SigLIP), WFS-SB performance is bounded by the semantic understanding capabilities of the underlying VLM. Poorly calibrated scores—due to domain shift, adversarial perturbations, or out-of-distribution content—may cause the wavelet analysis to miss meaningful semantic transitions. This underscores the importance of selecting an appropriate VLM for the target domain.

\noindent\textbf{Sensitivity to Extreme Temporal Structures.}
Videos with very rapid scene cuts (e.g., advertisements, montages) may trigger excessive boundary detection, leading to over-segmentation. Conversely, videos with extended low-relevance periods interrupted by brief high-relevance moments may cause the segment filtering mechanism to inadvertently discard important short segments. While our default hyperparameters ($\eta=1.2$) work well across diverse benchmarks, domain-specific tuning may be necessary for extreme cases.


\subsection{Future Work}
\label{sec:suppl_future_work}

Building on the identified limitations, we propose four promising research directions for extending wavelet-based frame selection.

\noindent\textbf{Efficient ITM Score Approximation.}
To address computational overhead, future work could explore: (1) \textit{Distillation-based approximation}, where a smaller model is trained to predict ITM scores from visual features; (2) \textit{Adaptive FPS sampling}, using lower frame rates for longer videos to reduce preprocessing time while maintaining performance; (3) \textit{Sparse ITM computation}, where only a subset of frames are scored initially, and additional frames are queried adaptively based on wavelet analysis. These strategies could significantly reduce preprocessing time.

\noindent\textbf{Learned Wavelet Kernels for Video Decomposition.}
While Table~\ref{tab:wavelet_ablation} demonstrates robustness across different wavelet families, all tested wavelets are hand-designed. An intriguing direction is learning \textit{data-driven wavelet bases} optimized for video semantic boundary detection through end-to-end training on video segmentation datasets, where wavelet filters are parameterized as learnable convolutional kernels. This could capture domain-specific temporal patterns more effectively than generic wavelets.


\noindent\textbf{Multi-Query and Open-Ended Video Understanding.}
Our current formulation assumes a single query per video. Future work could extend WFS-SB to \textit{multi-query scenarios} by jointly analyzing temporal relevance signals for all queries and selecting a unified frame set that maximizes coverage. For open-ended video understanding without explicit queries, the framework could be adapted to select keyframes based on \textit{self-supervised semantic change detection} using frame-to-frame similarity or anomaly detection.

\noindent\textbf{Extension to Multimodal Signals.}
While our focus is visual-textual alignment, many videos contain rich audio and subtitle information. Future work could extend wavelet-based boundary detection to \textit{multimodal temporal signals}, fusing audio energy, speech transcripts, and visual ITM scores into a unified relevance signal. This could enable more robust boundary detection in scenarios where visual cues alone are ambiguous (e.g., dialogue-heavy scenes where semantic transitions are signaled primarily by speech).

\section{Further Explanation of Our Method}
\label{sec:further method}
\label{sec:suppl_peak_detection}

Section~\ref{sec:wavelet_decomposition} briefly described the peak detection process for identifying semantic boundaries in the change intensity signal $c_t = |\tilde{s}_t|$. Here we provide the complete mathematical formulation and algorithmic pseudocode.

The algorithm identifies local maxima in $c_t$ that satisfy two adaptive criteria: \textit{height} and \textit{prominence}. These criteria ensure detected peaks correspond to genuine semantic boundaries rather than noise-induced fluctuations.

\noindent\textbf{Change Intensity Signal.}
Given the wavelet-reconstructed semantic change signal $\tilde{s}_t$ = $\text{IDWT}(\{\mathbf{0}, d_J, \mathbf{0}, \dots, \mathbf{0}\})$ (Equation~\ref{eq:wavelet_reconstruction}), we compute its absolute value to obtain a non-negative change intensity:
\begin{equation}
c_t = |\tilde{s}_t|, \quad t = 1, \dots, N.
\end{equation}
The peaks in $c_t$ correspond to moments of maximal semantic transition.

\noindent\textbf{Adaptive Height Threshold.}
To filter low-magnitude fluctuations, we require peaks to exceed a data-driven threshold:
\begin{equation}
\tau_{\text{height}} = \bar{c} + \alpha \cdot \sigma_c,
\end{equation}
where $\bar{c}$ is mean change intensity, $\sigma_c$ is standard deviation, and $\alpha$ is the height factor (default: $\alpha=0.5$). A peak at index $t$ is retained only if $c_t \geq \tau_{\text{height}}$.

\noindent\textbf{Adaptive Prominence Threshold.}
Prominence measures how much a peak stands out relative to surrounding valleys. 
We require prominence to exceed a threshold proportional to the signal's dynamic range:
\begin{equation}
\tau_{\text{prom}} = \beta \cdot (\max_t c_t - \min_t c_t),
\end{equation}
where $\beta$ is the prominence factor (default: $\beta=0.05$). This suppresses broad, low-contrast humps.

\noindent\textbf{Minimum Distance Constraint.}
To prevent over-segmentation, we enforce minimum temporal separation $\delta_{\text{min}}$ between consecutive peaks via non-maximum suppression. The minimum distance is adaptively set as:
\begin{equation}
\delta_{\text{min}} = \max\left(5, \left\lfloor N \times 0.02 \right\rfloor \right),
\end{equation}
ensuring at least 5 frames separation for short videos and proportionally larger separation for long videos.

Algorithm~\ref{alg:peak_detection} provides the complete procedural description of the peak detection process.

\begin{algorithm}[t]
\small
\caption{Adaptive Peak Detection for Semantic Boundaries}
\label{alg:peak_detection}
\begin{algorithmic}[1]
  \Require Change intensity signal $\{c_t\}_{t=1}^N$, height factor $\alpha$, prominence factor $\beta$
  \Ensure Set of boundary indices $\mathcal{B} = \{b_1, \dots, b_M\}$

  \State Compute signal statistics:
  \State \hspace{\algorithmicindent} $\bar{c} \gets \frac{1}{N} \sum_{t=1}^N c_t$ \Comment{Mean intensity}
  \State \hspace{\algorithmicindent} $\sigma_c \gets \sqrt{\frac{1}{N} \sum_{t=1}^N (c_t - \bar{c})^2}$ \Comment{Std deviation}
  \State \hspace{\algorithmicindent} $R_c \gets \max_t c_t - \min_t c_t$ \Comment{Dynamic range}

  \State Compute adaptive thresholds:
  \State \hspace{\algorithmicindent} $\tau_{\text{height}} \gets \bar{c} + \alpha \cdot \sigma_c$
  \State \hspace{\algorithmicindent} $\tau_{\text{prom}} \gets \beta \cdot R_c$
  \State \hspace{\algorithmicindent} $\delta_{\text{min}} \gets \max(5, \lfloor N \times 0.02 \rfloor)$

  \State Initialize candidate peak set: $\mathcal{P}_{\text{cand}} \gets \emptyset$

  \For{$t = 2$ \textbf{to} $N-1$}
    \If{$c_t > c_{t-1}$ \textbf{and} $c_t > c_{t+1}$} \Comment{Local maximum}
      \If{$c_t \geq \tau_{\text{height}}$}
        \State Compute prominence $p_t$ of peak at $t$
        \If{$p_t \geq \tau_{\text{prom}}$}
          \State $\mathcal{P}_{\text{cand}} \gets \mathcal{P}_{\text{cand}} \cup \{t\}$
        \EndIf
      \EndIf
    \EndIf
  \EndFor

  \State Sort $\mathcal{P}_{\text{cand}}$ by prominence in descending order
  \State Initialize final boundary set: $\mathcal{B} \gets \emptyset$

  \For{each peak $t_p \in \mathcal{P}_{\text{cand}}$ (in sorted order)}
    \If{$\mathcal{B} = \emptyset$ \textbf{or} $\min_{b \in \mathcal{B}} |t_p - b| \geq \delta_{\text{min}}$}
      \State $\mathcal{B} \gets \mathcal{B} \cup \{t_p\}$
    \EndIf
  \EndFor

  \State Sort $\mathcal{B}$ in ascending order
  \State \Return $\mathcal{B}$
\end{algorithmic}
\end{algorithm}

\noindent\textbf{Hyperparameter Robustness Analysis.}
\label{sec:suppl_peak_params}
Table~\ref{tab:peak_hyperparameter} evaluates robustness to the height factor $\alpha$ and prominence factor $\beta$ on VideoMME and MLVU with Qwen2.5-VL-7B ($K$=16).

\begin{table}[ht]
\centering
\small
\caption{Robustness of peak detection to hyperparameter variations. Evaluated on VideoMME and MLVU with Qwen2.5-VL-7B ($K$=16). Performance varies by only 1.1\% or 1.2\% across different settings, demonstrating strong hyperparameter insensitivity.}
\label{tab:peak_hyperparameter}
\begin{tabular}{cccc}
\toprule
$\alpha$ & $\boldsymbol{\beta}$ & \textbf{VideoMME} & \textbf{MLVU}\\
\midrule
\hl 0.5 &\hl 0.05 &\hl  61.9 & \hl 67.9\\
 0.0 & 0.05 &\textbf{62.6} &67.2 \\
 1.0 & 0.05 & 61.5&\textbf{68.4}\\
 0.5 & 0.00 & 61.8&67.6 \\
 0.5 & 0.1 & 62.0&67.9 \\
\bottomrule
\end{tabular}
\end{table}

The results demonstrate strong robustness: performance varies by only 1.1\% or 1.2\% across different hyperparameter settings. This insensitivity to $\alpha$ and $\beta$ indicates that the adaptive thresholding mechanism naturally adjusts to signal characteristics, making the method reliable across diverse video content without requiring task-specific tuning.

\section{Additional Experimental Results}
\label{sec:suppl_additional_experiments}

This section presents extended experimental results that complement the analyses in the main paper, including comparisons with additional baseline methods, extended frame budget and model ablations, performance under different frame rates, hyperparameter sensitivity studies, and qualitative visualizations.

\subsection{Comparison with More Baseline Methods}
\label{sec:suppl_more_baselines}

Table~\ref{tab:suppl_more_baselines} extends the main comparison (Table~\ref{tab:main_results}) by evaluating additional recent methods on Qwen2-VL-7B.

\begin{table*}[t]
\centering
\small
\setlength{\tabcolsep}{3pt}
\caption{Extended comparison with additional baseline methods on VideoMME, MLVU, and LongVideoBench using Qwen2-VL-7B ($K$=8). WFS-SB achieves competitive performance with consistent improvements across all benchmarks (+4.4\%, +8.4\%, +3.8\%). $^*$Same token budget but more frames.}
\label{tab:suppl_more_baselines}
\setlength{\tabcolsep}{3pt}  
\begin{tabular}{l|l|c|c|ccc|ccc|ccc}
\toprule
\multirow{2}{*}{\textbf{Model}} & \multirow{2}{*}{\textbf{Method}} & \multirow{2}{*}{\textbf{Size}} & \multirow{2}{*}{\textbf{Frame}} & \multicolumn{3}{c|}{\textbf{VideoMME}~\cite{fu2025videomme}} & \multicolumn{3}{c|}{\textbf{MLVU}~\cite{zhou2024mlvu}} & \multicolumn{3}{c}{\textbf{LongVideoBench}~\cite{wu2024longvideobench}} \\
\cmidrule(lr){5-7} \cmidrule(lr){8-10} \cmidrule(lr){11-13}
& & & & Base & +Method & $\Delta$ & Base & +Method & $\Delta$ & Base & +Method & $\Delta$ \\
\midrule

\multirow{3}{*}{Qwen2-VL-7B~\cite{wang2024qwen2vl}}
& KFC~\cite{fang2025threading} & 7B & 8 & 55 & 56.7 & \gain{+1.7} & 59.6 & \textbf{65.9} & \gain{+6.3} & 53.4 & 54.6 & \gain{+1.2} \\
& Q-Frame~\cite{zhang2025qframe} & 7B & 8 (4+8+32)$^*$ & 53.7 & \textbf{58.3} & \gain{\textbf{+4.6}} & 56.9 & 65.4 & \gain{\textbf{+8.5}} & 53.5 & \textbf{58.4} & \gain{\textbf{+4.9}} \\
& \hl WFS-SB & \hl 7B & \hl 8 & \hl 52.0 & \hl 56.4 & \hl \gain{+4.4} & \hl 55.0 & \hl 63.4 & \hl \gain{+8.4} & \hl 51.7 & \hl 56.5 & \hl \gain{+3.8} \\

\bottomrule
\end{tabular}
\end{table*}

\subsection{Extended Frame Budget and Model Ablation}
\label{sec:suppl_extended_frame_budget}

\begin{figure}[t]
\centering
\includegraphics[width=\linewidth]{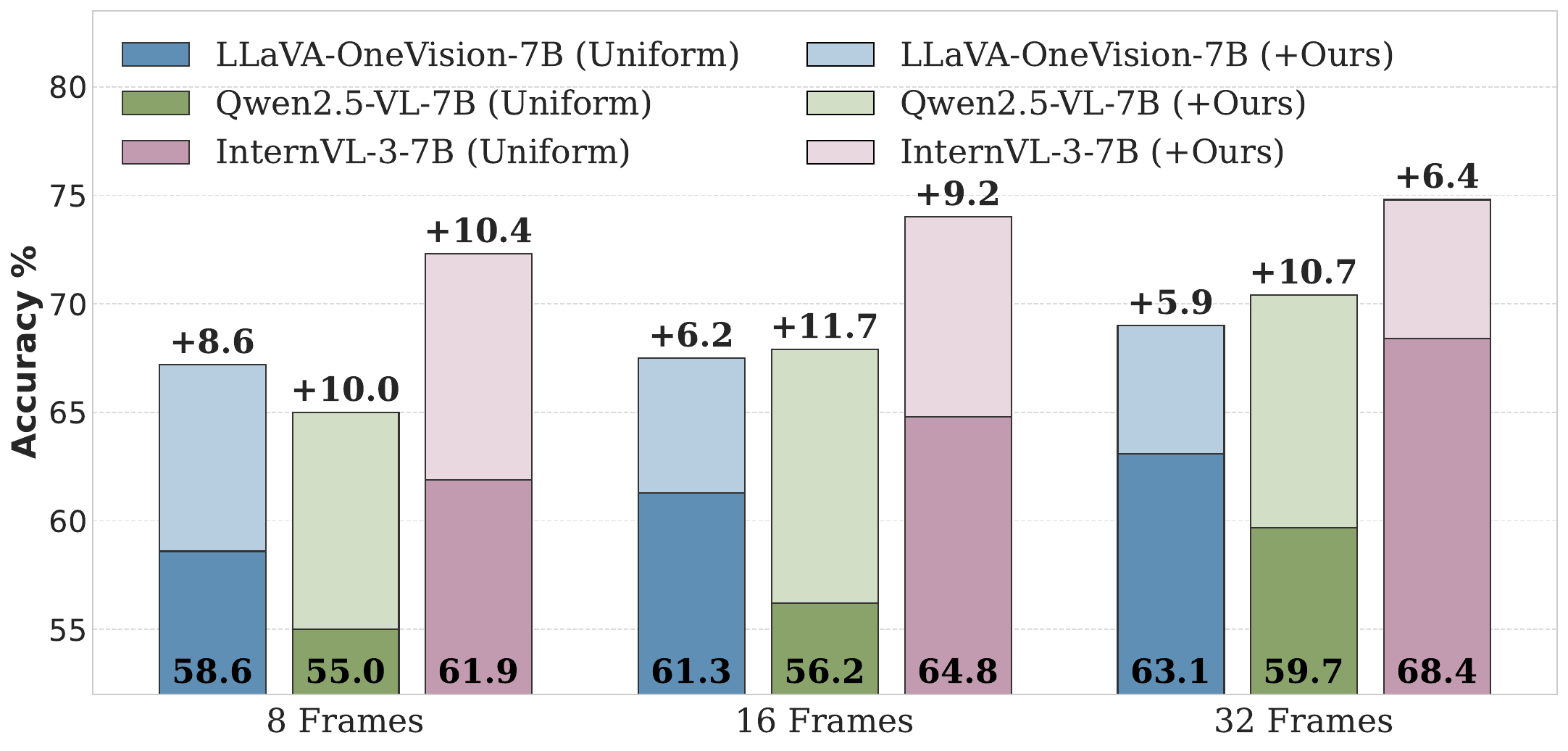}
\caption{
\textbf{Performance across different frame budgets on MLVU.} WFS-SB consistently outperforms uniform sampling across all LVLMs and budget settings.
}
\label{fig:mlvu_model_comparison}
\end{figure}

\begin{figure}[t]
\centering
\includegraphics[width=\linewidth]{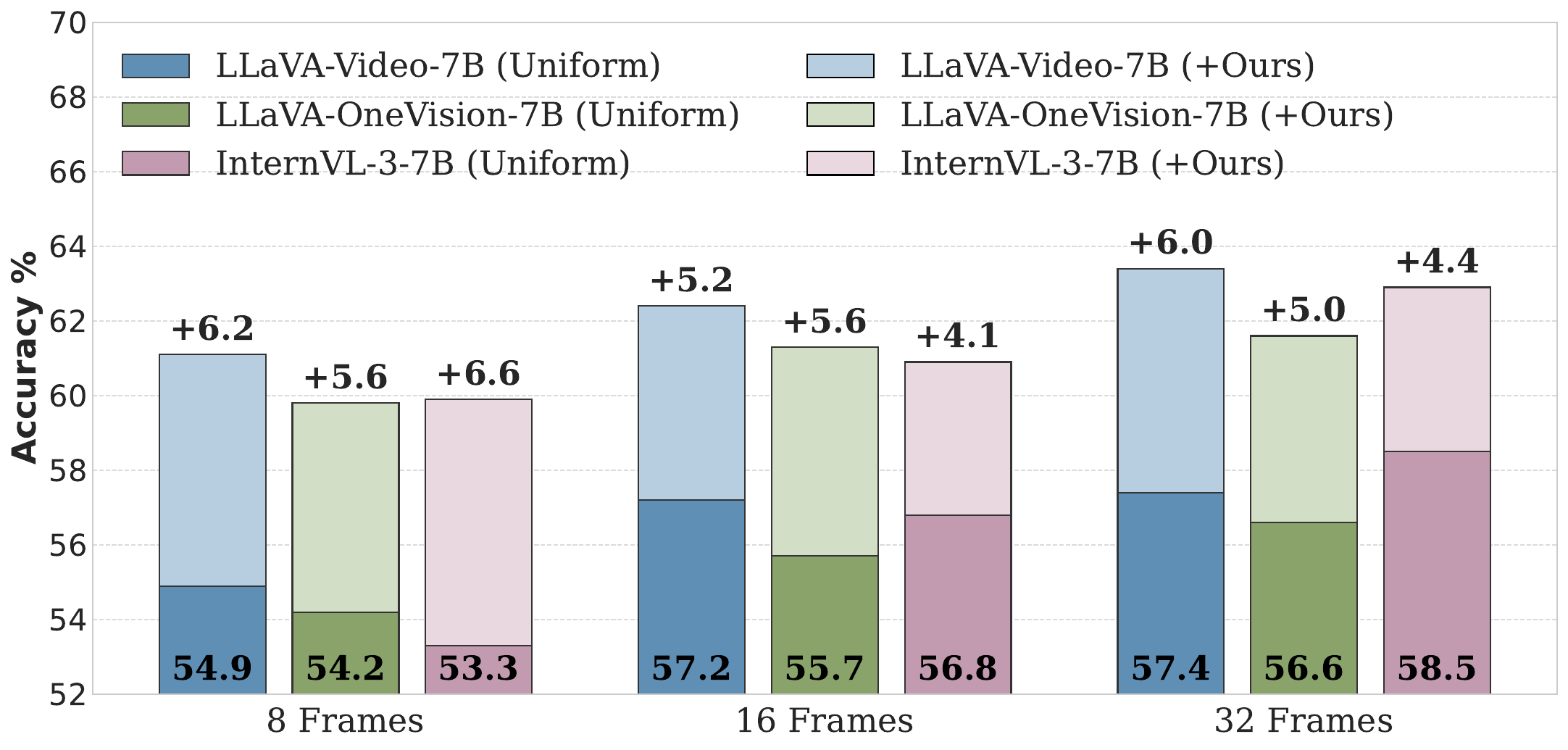}
\caption{
\textbf{Performance across different frame budgets on LongVideoBench.} WFS-SB demonstrates robust improvements across multiple architectures and budget constraints.
}
\label{fig:lvb_model_comparison}
\end{figure}

Figure~\ref{fig:mlvu_model_comparison} and Figure~\ref{fig:lvb_model_comparison} extend the frame budget analysis from Figure~\ref{fig:frame_budget} to MLVU and LongVideoBench. WFS-SB consistently outperforms uniform sampling across all models and budgets, with particularly strong gains at smaller budgets ($K$=8, 16).

\subsection{Performance Across Different Frame Rates}
\label{sec:suppl_fps_ablation}

Our default pipeline samples candidate frames at 1 FPS. To address the ITM extraction bottleneck (79\% of preprocessing time, Table~\ref{tab:efficiency}), we evaluate an adaptive FPS strategy that uses different sampling rates for videos of different durations in VideoMME. Table~\ref{tab:suppl_fps_ablation} reports results on VideoMME with Qwen2.5-VL-7B.

\begin{table}[ht]
\centering
\small
\setlength{\tabcolsep}{4pt}
\caption{Impact of adaptive FPS sampling on performance and ITM extraction time. Evaluated on VideoMME with Qwen2.5-VL-7B. The notation "1-0.75-0.5" indicates using 1 fps for short videos, 0.75 fps for medium videos, and 0.5 fps for long videos. Adaptive strategies reduce ITM time by 46\% (1-0.75-0.5) and 70\% (1-0.5-0.25) respectively while maintaining or improving performance. F. represents the number of input frames.}
\label{tab:suppl_fps_ablation}
\begin{tabular}{lcccc}
\toprule
\textbf{Sampling Rate} & \textbf{8 F.}  & \textbf{16 F.} & \textbf{32 F.} & \textbf{ITM Time (s)} \\
\midrule
 Uniform Sampling& 53.2 & 57.7 & 61.2 & - \\
 \hl 1 fps&\hl 59.3 &\hl 61.9 &\hl 64.4 &\hl 19.4 \\
1-0.75-0.5 fps& 58.9 & \textbf{62.0} & 64.5 & 10.5 \\
1-0.5-0.25 fps& \textbf{59.4} & 61.9 & \textbf{64.6} & 5.8 \\
\bottomrule
\end{tabular}
\end{table}

The results demonstrate that adaptive FPS strategies significantly reduce computational overhead. By using lower sampling rates for medium and long videos (e.g., 1-0.5-0.25 fps for short-medium-long videos), ITM extraction time decreases from 19.4s to 5.8s (70\% reduction) while maintaining comparable or even slightly improved performance. This validates that reducing FPS for longer videos is a practical approach to mitigate the preprocessing time bottleneck without sacrificing accuracy.

\subsection{Extended VLM Comparison for Query-Frame Matching}
\label{sec:suppl_extended_vlm}

Table~\ref{tab:more_vlm_comparison} extends the VLM comparison from Table~\ref{tab:vlm_comparison} by evaluating on LLaVA-Video-7B with $K$=16 across all three benchmarks.

\begin{table}[ht]
\centering
\small
\caption{Ablation on different VLMs for query-frame similarity scoring with LLaVA-Video-7B ($K$=16). We select BLIP-2-ITM as default for its superior performance across all benchmarks.}
\label{tab:more_vlm_comparison}
\begin{tabular}{lccc}
\toprule
\textbf{VLM Scorer} & \textbf{VideoMME} & \textbf{MLVU} & \textbf{LVB} \\
\midrule
Uniform & 60.6 & 60.9 & 57.2 \\
\midrule
BLIP-ITM~\cite{li2022blip}             & 63.2& 69.5 & 61.8 \\
CLIP-VIT-B~\cite{radford2021learning}              & 63.6 & 67.3 & 62.2 \\
SigLIP-so400m~\cite{zhai2023sigmoid}            &  62.4&  68.9&61.5  \\
\hl BLIP-2-ITM (Ours)~\cite{li2023blip2} & \hl \textbf{64.3} & \hl \textbf{71.0} & \hl \textbf{62.4} \\
\bottomrule
\end{tabular}
\end{table}

The results demonstrate WFS-SB's robustness to VLM choice, with all tested models yielding substantial improvements over uniform sampling. Although BLIP-ITM performs better under the settings in Table~\ref{tab:vlm_comparison}, we select BLIP-2-ITM as default because it provides more consistent average benefits across diverse settings. Specifically, BLIP-2-ITM achieves the best performance, with accuracies of 64.3\%, 71.0\%, and 62.4\% on VideoMME, MLVU, and LongVideoBench, respectively, demonstrating its superior cross-benchmark generalization ability. This reflects BLIP-2-ITM's better calibration from its enhanced image-text alignment objective.

\subsection{Hyperparameter Ablation: Segment Filtering}
\label{sec:suppl_segment_filtering}

Section~\ref{sec:budget_allocation} introduced a segment filtering mechanism controlled by threshold $\tau = \text{mean}(\text{Imp}) - \eta \cdot \text{std}(\text{Imp})$, where $\eta$ controls filtering aggressiveness. Table~\ref{tab:suppl_segment_filtering} analyzes the impact of $\eta$ on performance.

\begin{table}[ht]
\centering
\small
\caption{Robustness of segment filtering to hyperparameter $\eta$. Evaluated on VideoMME and MLVU with Qwen2.5-VL-7B. Performance varies by only 0.2-0.6\% across different $\eta$ values. Filtering is effective: compare "-" (no filtering, all segments retained) vs. others. }
\label{tab:suppl_segment_filtering}
\begin{tabular}{ccc}
\toprule
$\boldsymbol{\eta}$ & \textbf{VideoMME} & \textbf{MLVU}\\
\midrule
\hl 1.2 &\hl  61.9 & \hl 67.9\\
- & 61.5&67.2 \\
1.0& 61.7&67.5\\
1.5 & 61.9&68.1 \\
\bottomrule
\end{tabular}
\end{table}

The results demonstrate two key properties. \textbf{Robustness:} Performance varies by only 0.2-0.6\% across different $\eta$ values, indicating the method is insensitive to this hyperparameter. \textbf{Effectiveness:} Comparing the no-filtering baseline ("-") with filtered configurations shows that segment filtering consistently improves performance by 0.4-0.9\%, as it concentrates the frame budget on high-relevance segments while discarding low-importance regions. The default $\eta=1.2$ provides a reliable balance.

\subsection{Additional Qualitative Examples}
\label{sec:suppl_qualitative}

\begin{figure*}[t]
\centering
\includegraphics[width=\linewidth]{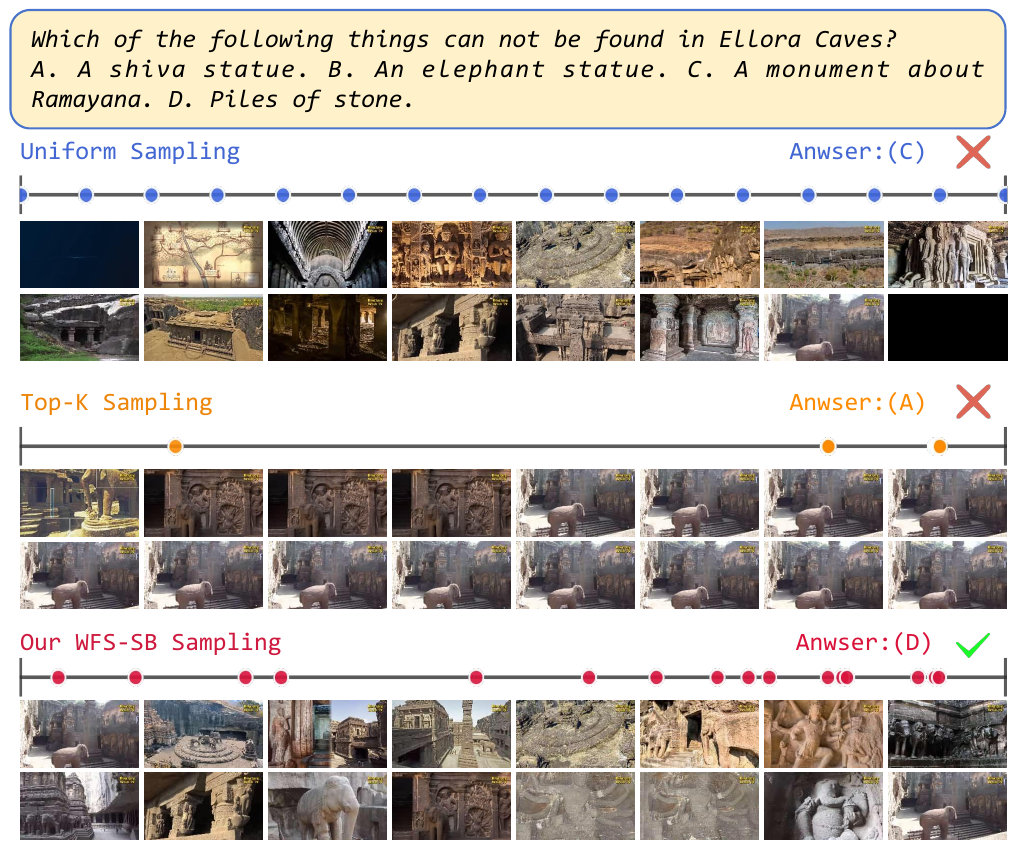}
\caption{
\textbf{Qualitative comparison on Ellora Caves video.} Query: \textit{"Which of the following things can not be found in Ellora Caves? A. A shiva statue. B. An elephant statue. C. A monument about Ramayana. D. Piles of stone."}
\textbf{Top:} Uniform sampling (Answer: C) - Too random, lacks query-specific awareness, missing critical frames.
\textbf{Middle:} Top-K sampling (Answer: A) - Focuses on highly salient objects (e.g., elephant statue) but misses other equally important yet less visually prominent frames.
\textbf{Bottom:} Our WFS-SB (Answer: D, correct) - Through wavelet-based boundary detection, effectively extracts semantic boundaries to segment the video. Adaptive budget allocation and intra-segment diversity selection successfully identify keyframes that enable correct question answering.
}
\label{fig:suppl_qual_example1}
\end{figure*}

Figure~\ref{fig:suppl_qual_example1} provides a qualitative comparison of different frame selection strategies on a video from VideoMME.

This example illustrates the advantages of the WFS-SB Method. Uniform sampling fails due to its random, query-agnostic nature, missing critical content. Top-K sampling exhibits bias toward visually salient objects (e.g., prominent elephant statue), overlooking other equally important but less conspicuous elements necessary for answering the question. In contrast, WFS-SB leverages wavelet-based semantic boundary detection to partition the video into coherent segments, then applies adaptive budget allocation and diversity-aware selection within segments to capture comprehensive coverage of all relevant content, enabling accurate question answering.



\end{document}